\pdfoutput=1

\documentclass[11pt]{article}

\usepackage[final]{coling}

\usepackage{times}
\usepackage{latexsym}

\usepackage[T1]{fontenc}

\usepackage[utf8]{inputenc}

\usepackage{microtype}

\usepackage{inconsolata}

\usepackage{multirow}
\usepackage{times}
\usepackage{latexsym}
\usepackage{amssymb}
\usepackage{amsmath}
\usepackage{graphicx}
\usepackage{booktabs}%
\usepackage{subfigure}
\usepackage{wrapfig}
\usepackage{dblfloatfix}
\title{Unifying Dual-Space Embedding for Entity Alignment via Contrastive Learning}

\author{
\textbf{Cunda Wang\textsuperscript{1}},
 \textbf{Weihua Wang\textsuperscript{1,2,3,}\thanks{Corresponding Author. Email: \href{mailto:wangwh@imu.edu.cn}{wangwh@imu.edu.cn.}}},
 \textbf{Qiuyu Liang\textsuperscript{1}},
 \textbf{Feilong Bao\textsuperscript{1,2,3}},
 \textbf{Guanglai Gao\textsuperscript{1,2,3}}
\\
\\
 \textsuperscript{1}College of Computer Science, Inner Mongolia University, Hohhot, China
 \\
 \textsuperscript{2}National and Local Joint Engineering Research Center of Intelligent 
 Information 
 \\
 Processing Technology for Mongolian, Hohhot, China
 \\
 \textsuperscript{3}Inner Mongolia Key Laboratory of Multilingual Artificial Intelligence Technology, Hohhot, China
\\
}

\begin{document}
\maketitle
\begin{abstract}
Entity alignment aims to match identical entities across different knowledge graphs (KGs).
Graph neural network-based entity alignment methods have achieved promising results in Euclidean space.
However, KGs often contain complex structures, including both local and hierarchical ones, which make it challenging to efficiently represent them within a single space.
In this paper, we proposed a novel method UniEA, which unifies dual-space embedding to preserve the intrinsic structure of KGs.
Specifically, we learn graph structure embedding in both Euclidean and hyperbolic spaces simultaneously to maximize the consistency between the embedding in both spaces. 
Moreover, we employ contrastive learning to mitigate the misalignment issues caused by similar entities, where embedding of similar neighboring entities within the KG become too close in distance.
Extensive experiments on benchmark datasets demonstrate that our method achieves state-of-the-art performance in structure-based EA. Our code is available at \url{https://github.com/wonderCS1213/UniEA}.
\end{abstract}

\section{Introduction}
Knowledge graphs (KGs) represent real-world knowledge in the form of graphs. 
They typically store data in the form of triples $(h,r,t)$, where $h$ represents the head entity, $r$ the relation, and $t$ the tail entity.
The completeness of KGs affects tasks such as knowledge-driven question answering \citep{qa} and recommendation \citep{DBLP:conf/iclr/Cai0XR23, qyrec}. 
Hence, it is essential to integrate multiple source KGs to build a comprehensive KG.
Entity alignment (EA) serves as an important step in this process. 
It aims to identify the same real-world entities referenced across different KGs.
\begin{figure}[t]
  \includegraphics[width=\columnwidth]{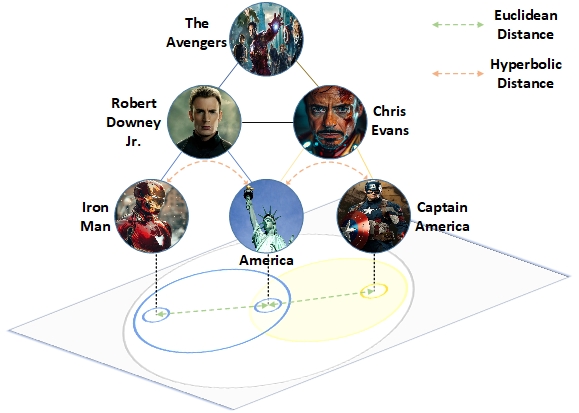}
  \caption{Knowledge graph with hierarchical structures.}
  \label{fig:des}
\end{figure}

Recently, due to the strong neighborhood learning capabilities of graph neural networks (GNNs), GNN-based EA have achieved significant progress\citep{Xie2023ImprovingKG,gsea,Sun2019KnowledgeGA}.
However, GNNs face two issues in Euclidean space embedding: 1) limited performance when handling complex hierarchical structures, and 2) the embeddings of neighboring entities are overly similar.

As shown in Figure~\ref{fig:des}, this is a common type of hierarchical structure found in KGs.
Traditional GNN-based EA methods often embed entities like ``Iron Man'' and ``America'' directly according to their Euclidean distance. Nevertheless, this does not reflect the true distance between these two entities, leading to distortion in the graph structure embeddings.
The hyperbolic space can capture the hierarchical structure of graphs \citep{wangwh, QY2}. The hyperbolic distance better represents the true distance between the entities ``Iron Man'' and ``America''.
Moreover, these methods \citep{Wang2018CrosslingualKG, ke-gcn} cause similar entities within the same KG to have embeddings that are too close in distance.
For example, entities like ``Robert Downey Jr.'' and ``Chris Evans'' share multiple neighboring entities, such as ``The Avengers`` and ``America``. These shared neighbors often lead to homogenization, resulting in incorrect entity alignment.
Current methods have proposed various solutions to these two challenges \citep{Sun2020KnowledgeAW,HMEA,Xie2023ImprovingKG,gsea}. For instance, \citet{Sun2020KnowledgeAW} and \citet{HMEA} explore EA task in hyperbolic space embedding, demonstrating that hyperbolic space is more effective for learning the hierarchical structure of graphs, which aids in entity alignment.
\citet{Xie2023ImprovingKG} alleviates over-smoothing through graph augmentation techniques.
However, the augmentation strategies, which randomly perturb the graph topology, may degrade the quality of the graph embeddings \citep{aug}. 
Our motivation is to consider hyperbolic space embedding as an augmentation of graph embedding. 
This approach not only avoids the drawbacks of traditional graph augmentation techniques but also leverages the hierarchical structure information provided by hyperbolic embedding.

To address the aforementioned issues, we propose a novel method named UniEA, which \textbf{Uni}fies the Euclidean and hyperbolic spaces embedding for \textbf{EA}.
Our method is not limited to embedding in a single space.
Specifically, we introduce graph attention networks (GATs) \citep{gat} to aggregate neighboring entities in Euclidean space and employ hyperbolic graph convolutional networks (HGCNs) \citep{hgcn} to learn the hierarchical structural information of the graph in hyperbolic space.
We maximize the consistency between the embedding in Euclidean space and hyperbolic space through \textbf{contrastive learning}, which leads to more accurate entity embeddings.
Moreover, the close distances of similar neighboring embedding severely affect the final alignment of entities. We employ \textbf{contrastive learning once again} to address the issue.
The contributions of this work can be summarized as follows:
\begin{itemize}
    \item We propose a novel EA method called UniEA. To the best of our knowledge, this is the first method for EA that leverages contrastive learning to unify Euclidean and hyperbolic space embeddings.
    \item We also employ contrastive learning to mitigate misalignment issues caused by overly close distances between similar entity embeddings.
    \item The extensive experiments on four public datasets demonstrate that UniEA consistently outperforms the state-of-the-art methods for structure-based EA.
\end{itemize}

\section{Related work}
In line with our work, we review related work in three areas: EA in Euclidean space, representation learning in hyperbolic space and improving EA with graph augmentation.
\subsection{EA in Euclidean Space}
Current embedding-based EA methods can be broadly categorized into three types: TransE-based EA, GNN-based EA and other methods. All of these primarily aim to learn embeddings for entities and relations from relational triples.

Due to the strong performance of TransE \citep{TransE} in capturing local semantic information of entities, several methods have proposed variants of TransE for application in EA.
For instance, 
\citet{Chen2016MultilingualKG} addresses the inconsistency in cross-lingual embedding spaces. 
\citet{Zhu2017IterativeEA} emphasizes path information. 
\citet{Sun2018BootstrappingEA} treats EA as a classification task. \citet{Pei2019SemiSupervisedEA} enhances knowledge graph embedding by leveraging nodes with varying degrees.

TransE-based EA methods lack the ability to effectively model global structural information. As a result, recent research increasingly favors GNN-based approaches for EA.
Stacking multiple layers of GNNs enables the capture of information from more distant neighbors, which facilitates learning of global structural information.
For example, \citet{Wang2018CrosslingualKG} directly stacks multiple layers of vanilla GCN \citep{gcn} to obtain entity embeddings.
Due to the heterogeneity of KGs, the alignment performance is limited.  
\citet{Sun2019KnowledgeGA} employs a gating mechanism to attempt capturing effective information from distant neighbors. MRAEA \citep{Mao2020MRAEAAE}, RAEA \citep{Zhu2021RAGARG}, KE-GCN \citep{ke-gcn}, RSN4EA \citep{DBLP:journals/corr/abs-1905-04914}, GAEA \citep{Xie2023ImprovingKG}, RHGN \citep{Liu2023RHGNRH} , and GSEA \citep{gsea} utilize rich relational information to obtain entity embeddings.
\citet{Xin2022InformedME} encoded neighbor nodes, triples, and relation paths together with transformers.
Unfortunately, the ability to handle complex topological structures in graphs is limited in Euclidean space.

Additionally, some methods integrate the rich information within KGs to enhance the performance of EA tasks. This includes leveraging attributes \citep{attrgnn}, entity names \citep{bertint} and more \citep{Meaformer}. \citet{chatEA} explores the potential of large language models for EA task.
Since our method focuses on structural information, we do not compare it with the above methods to ensure experimental fairness.
\subsection{Representation learning in hyperbolic space}
Hyperbolic space has recently garnered considerable attention due to its strong potential for learning hierarchical structures and scale-free characteristics.
For example,
\citet{hgcn} first introduced the use of graph convolutional networks (GCNs) and hyperbolic geometry through an inductive hyperbolic GCN.

Hyperbolic space representation learning has been applied to various downstream tasks, achieving excellent performance in areas such as node classification \citep{QY2}  and completion \citep{QY3, qynlpcc}. Notably, existing work has successfully completed EA using hyperbolic space embedding.
\citet{Sun2020KnowledgeAW} extends translational and GNN-based techniques to hyperbolic space, and captures associations by a hyperbolic transformation. 
\citet{HMEA} integrates multi-modal information in the hyperbolic space and predict the alignment results based on the hyperbolic distance.
Although these methods demonstrate the advantages of hyperbolic embedding, they are limited to embedding solely in hyperbolic space.
\subsection{Improving EA with graph augmentation}
Graph augmentation techniques primarily generate augmented graphs by perturbing the original graph through node dropout or edge disturbance, effectively enhancing the model's robustness to graph data.

Graph augmentation techniques have been proven effective in entity alignment tasks. GAEA \citep{Xie2023ImprovingKG} opts to generate augmented graphs by removing edges rather than adding new ones, as introducing additional edges can lead to extra noise. GSEA \citep{gsea} employs singular value decomposition to generate augmented graphs, capturing the global structural information of the graph. It leverages contrastive loss to learn the mutual information between the global and local structures of entities. However, these methods fall short in effectively learning the hierarchical structure of graphs.

\section{Preliminaries}
In this section, we define the EA task and explain the fundamental principles of hyperbolic space. This foundation is essential for comprehending our approach.
\subsection{Entity alignment}
Formally, we repesent a KG as $ \mathcal{G}=\{\mathcal{E},\mathcal{R},\mathcal{T}\} $, where $\mathcal{E}$ denotes entities, $ \mathcal{R} $ denotes relations, $ \mathcal{T}=\mathcal{E}\times \mathcal{R}\times \mathcal{E} $ repesents triples.
Given two KGs, $ \mathcal{G}_1=\{\mathcal{E}_1,\mathcal{R}_1,\mathcal{T}_1\}  $ repesent source KG, $ \mathcal{G}_2=\{\mathcal{E}_2,\mathcal{R}_2,\mathcal{T}_2\}  $ repesent target KG.
EA aims to discern each entity pair$(e^1_i,e^2_i)$, $
e^1_i\in \mathcal{E}_1$, $e^2_i\in \mathcal{E}_2$ where $ e^1_i $ and $ e^2_i $ correspond to an identical real-world entity $e_i$.
Typically, we use pre-aligned seed entities $ \mathcal{S} $ to unify the embedding spaces of two KGs in order to predict the unaligned entities.
\subsection{Hyperbolic space}
Hyperbolic geometry is a non-Euclidean geometry with a constant negative curvature, where curvature measures how a geometric object deviates from a flat plane \citep{lwodim}.
Here, we use the $d$-dimensional Poincaré ball model with negative curvature $-c \space (c>0): H^{(d,c)}=\{x\in R^d: \parallel x \parallel^2 < \frac{1}{c} \} $.
For each point $ x \in H^{(d, c)} $, the tangent space (a sub-space of the Euclidean space) $ T_xH_c $ is a $d$-dimensional vector space at point $ x $, which contains all possible directions of path in $ H^{(d,c)} $ leaving from $x$. Then, we introduce two basic operations that exponential and logarithmic maps in the hyperbolic space.

Let $ \alpha  $ be the feature vector in the tangent space $ T_oH_c $; $ o $ is a point in the hyperbolic space $ H^{(d,c)} $, which is also as a reference point. Let $ o $ be the origin, $ o = 0$, the tangent space $ T_oH_c $ can be mapped to $H^{(d,c)}$ via the exponential map:
\begin{equation} \label{exp}
\exp_{o}^c(\alpha )=tanh(\sqrt{c}\|\alpha \|)\frac{\alpha }{\sqrt{c}\|\alpha \|}.
\end{equation}

Conversely, the logarithmic map which maps $\beta $ to $  T_oH_c $ is defined as:
\begin{equation} \label{log}
    \log_{o}^c(\beta  )=arctanh(\sqrt{c}\|\beta  \|)\frac{\beta  }{\sqrt{c}\|\beta  \|}.
\end{equation}
Here, $ \beta $ is hyperbolic space embedding.
\section{Method}
\begin{figure*}[ht]
  \includegraphics[width=\textwidth]{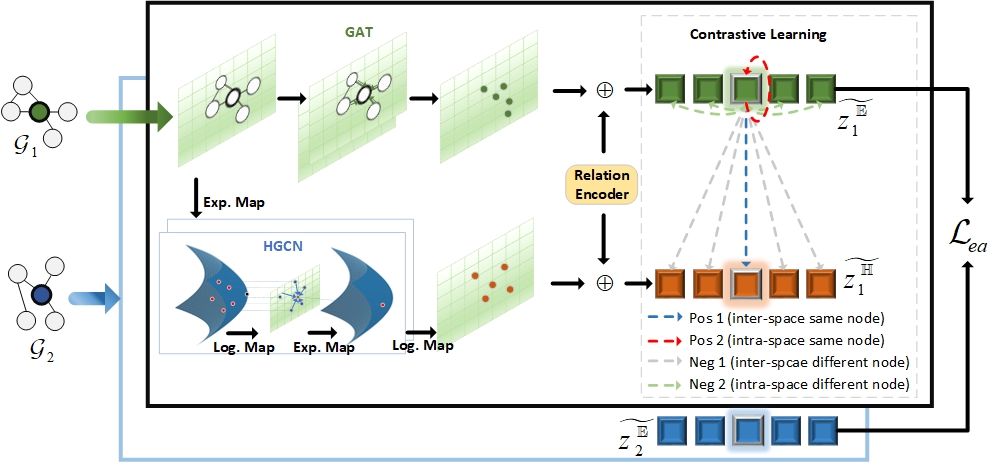}
  \caption{The framework of our proposed UniEA. Here, $ \oplus $ denotes concatenate. The `Exp. Map' operation is derived from Equation~\ref{exp}; the `Log. Map' operation is derived from Equation~\ref{log}.}
  \label{fig:HECL}
\end{figure*}

In this section, we elaborate on our approach in four parts.
As shown in Figure~\ref{fig:HECL}, our method includes: 1) Euclidean space embedding, 2) hyperbolic space embedding, 3) relation encoding and fusion, and 4) the loss function.

We randomly initialize the entity and relation embedding of $\mathcal{G}_1$, represented as $\textbf{z}_1^{\mathbb{E}} \in \mathbb{R}^{\mid \mathcal{E}_1 \mid \times d_e}$ and $\textbf{r}_1\in \mathbb{R}^{\mid \mathcal{R}_1 \mid \times d_r}$, respectively. Similarly, the entity and relation embedding of $\mathcal{G}_2$ are represented as $\textbf{z}_2^{\mathbb{E}}\in \mathbb{R}^{\mid \mathcal{E}_2 \mid \times d_e}$ and $\textbf{r}_2\in \mathbb{R}^{\mid \mathcal{R}_2 \mid \times d_e}$.
Here, $ \textbf{z}^{\mathbb{E}} $ denotes Euclidean space embedding; $d_e$ and $d_r$ stand for the dimensionality of entity and relation, respectively.
\subsection{Euclidean space embedding}
The ability of GAT to aggregate neighbor information in heterogeneous graphs has been well demonstrated \citep{Meaformer,gsea}. We stack multiple layers of GAT to obtain Euclidean space embedding:
\begin{equation}
\begin{aligned}
    \textbf{Z}^{\mathbb{E}} &=  [\mathbf{z}^{(1)},...,\mathbf{z}^{(L)}] \\
    &= GAT(\textbf{W}_{m},\textbf{M},\textbf{z}^{\mathbb{E},0}),
\end{aligned}
\end{equation}
where $ \textbf{M} $ denotes the graph adjacency matrix, $\textbf{W}_m \in \mathbb{R}^{d\times d}$ is a diagonal weight matrix for linear transformation.

Due to the varying importance of the neighborhoods aggregated by different layers of GAT.
For example, in Figure~\ref{fig:des}, aggregating the first-order neighbors of ``Chris Evans'' is most beneficial. While aggregating higher-order neighbors can capture some implicit relationships of the entity, it often introduces noise.
Therefore, \citet{Xie2023ImprovingKG} introduce an attention mechanism \citep{Vaswani2017AttentionIA} to assign different weights to the embeddings obtained from different layers:
\begin{equation}
\begin{aligned}
&[\hat{\mathbf{z}}^{(1)},...,\hat{\mathbf{z}}^{(L)}]\\
&=\mathrm{softmax}(\frac{(\mathbf{Z}^{\mathbb{E}}\mathbf{W}_q)(\mathbf{Z}^{\mathbb{E}}\mathbf{W}_k)^\top}{\sqrt{d_{e}}})\mathbf{Z}^{\mathbb{E}},
\end{aligned}
\end{equation}
where $ 1/\sqrt{d_e} $ is the scaling factor, $\mathbf{W}_q  $ and $\mathbf{W}_k$ are the learnable paramenter matrices. Finally, the Euclidean space embedding $\overline{\mathbf{z}^{\mathbb{E}}} = \frac1L\sum_{l=1}^L\hat{\mathbf{z}}^{(l)} $.

\subsection{Hyperbolic Space embedding}
Our method equips HGCN \citep{hgcn} to learn the hierarchical structure of graphs in hyperbolic space.

Specifically, we project Euclidean space embeddings $ \mathbf{z}^{\mathbb{E}} $ to hyperbolic space using exponential map (Equation \ref{exp}):
\begin{equation}
     \mathbf{z}^{\mathbb{H}} = \exp^c_o(\mathbf{z}^{\mathbb{E}}),
\end{equation}
where $\mathbf{z}^{\mathbb{H}} \in H^{(d,c)}$, in other words, we obtain the first layer of embedding $ \mathbf{z}^{\mathbb{H},{0}}$ in hyperbolic space.

For the hyperbolic space embedding of the $l$-th layer, by hyperbolic feature aggregation, we can get the hyperbolic embedding of the next layer.
The hyperbolic aggregation process is as follows:
\begin{equation}
\label{agg}
    \mathbf{z}^{\mathbb{H},{l+1}}=\exp_o^{c}(\sigma({\textbf{A}} \log_o^{c}(\mathbf{z}^{\mathbb{H},{l}})\textbf{W}_{l})).
\end{equation}
\textbf{A} represents the symmetric normalized adjacency matrix, $ \sigma $ is $ReLU(\cdot)$ and $\textbf{W}_l$ is a trainable weight matrix.

For example, for the input $ \mathbf{z}^{\mathbb{H},{0}}$ in $0$-th layer, we can get $ \mathbf{z}^{\mathbb{H},{1}}$ using Equation \ref{agg}.

Finally, we can obtain the final output $\mathbf{z}^{\mathbb{H},{L}}$ in Hyperbolic Space.
The `$L$' is a hyper-parameter denoting the number of layers of the HGCN.

\subsection{Relation encoding and fusion}
The same entities often share similar relations, and relational semantic information is also highly beneficial for EA.
\citet{Mao2020MRAEAAE} reveals that relying solely on the inflow direction to accumulate neighboring information through directed edges is insufficient. Accumulating information from the outflow direction as well would be highly beneficial.
This idea facilitates the bridging and propagation of more information in such a sparse graph.
Hence, following this work, we use both in-degree and out-degree relation encoders to learn the representation of relations:
\begin{equation}
\overline{r}_{e_i}=\frac{\textbf{A}_{e_i}^{rel_{in}}\textbf{r}}{|N_{{e_i}}^{in}|}\oplus\frac{\textbf{A}_{e_i}^{rel_{out}}\textbf{r}}{|N_{{e_i}}^{out}|},
\end{equation}
where $ |N_{e_i}^{in}| $ and $|N_{e_i}^{out}|$ are the in-degree and out-degree of $e_i$, respectively. $\textbf{A}^{rel_{in}}$ denotes the adjacency matrix for in-degrees, $ \textbf{r} $ represents relation embedding.

Please note that before fusion, the hyperbolic space embedding are projected to Euclidean space $ \overline{\mathbf{z}^{\mathbb{H}}} = \log_o^{L}(\mathbf{z}^{\mathbb{H},{L}}) $.
Through the steps above, we concatenate the entity-level and relation-level features in Euclidean space to obtain the final output. 
\begin{equation}
    \tilde{\mathbf{z}^{\mathbb{H}}} =\overline{\mathbf{z}^{\mathbb{H}}} \oplus  \overline{r}, \tilde{\mathbf{z}^{\mathbb{E}}} =\overline{\mathbf{z}^{\mathbb{E}}} \oplus  \overline{r}.
\end{equation}
Here, $\tilde{\mathbf{z}^{\mathbb{H}}}$ and $\tilde{\mathbf{z}^{\mathbb{E}}}$ denote final embedding in  hyperbolic space and Euclidean space, respectively.
\subsection{Loss function}
Our loss function consists of three components: (i) a contrastive loss for aligning Euclidean and hyperbolic space embeddings $ \mathcal{L}_{inter} $, (ii) an intra-graph contrastive loss to mitigate the issue of neighboring entity embeddings being too similar $ \mathcal{L}_{intra} $, and (iii) a margin-based alignment loss for the entity alignment task $ \mathcal{L}_{ea} $.
\subsubsection{Contrastive learning}
To ensure that the Euclidean space embedding retain their structure without distortion, we first use contrastive learning $ \mathcal{L}_{inter} $ to maximize the consistency \citep{Xie2023ImprovingKG, aug} between the Euclidean and hyperbolic space embedding.
Moreover, previous methods suggest that similar entity embedding should be closer \citep{cl}, but being too close can negatively affect the results of EA.
Therefore, we employ contrastive learning $ \mathcal{L}_{intra} $, aiming to push the distances between all entities within a graph further apart. We define the contrastive learning formula as follows:
\begin{equation} \label{lc}
   \mathcal{L}_{c,i}^{(G^\mathbb{E},G^{\mathbb{H}})}=-\mathrm{log}\frac{\exp(\langle \tilde{\mathbf{z}^{\mathbb{E}}_i},\tilde{\mathbf{z}^{\mathbb{H}}_i\rangle)}} {\sum_{k\in \mathcal{E}}\exp(\langle \tilde{\mathbf{z}^{\mathbb{E}}_i},\tilde{\mathbf{z}^{\mathbb{H}}_k}\rangle)},
\end{equation}
\begin{equation}
     \mathcal{L}_{inter}=\sum_{n=\{1,2\}}\frac{1}{2|\mathcal{E}_n|}\sum_{i\in \mathcal{E}_n}(\mathcal{L}_{c,i}^{(\tilde{G^{\mathbb{E}}_n},\tilde{G^{\mathbb{H}}_n})}+\mathcal{L}_{c,i}^{(\tilde{G^{\mathbb{H}}_n},\tilde{G^{\mathbb{E}}_n})}),
\end{equation}
\begin{equation}
    \mathcal{L}_{intra}=\sum_{i\in \mathcal{E}}\mathcal{L}_{c,i}^{(\tilde{G^{\mathbb{E}}_1},\tilde{G^{\mathbb{E}}_1})}
\end{equation}
Here, $ \mathcal{L}_{intra} $ and $ \mathcal{L}_{inter} $ can be calculated using the Equation~\ref{lc}. 
\subsubsection{Margin-based alignment loss}
We use the pre-aligned entity pairs $\mathcal{S}$ to bring the embeddings of the same entities in $\mathcal{G}_1$ and $\mathcal{G}_2$ closer, while pushing the embeddings of different entities further apart.
We choose to use Euclidean space embedding for the margin-based alignment loss:
\begin{equation}
\begin{aligned}
      \mathcal{L}_{ea}= & \sum_{(e_i,e_j)\in S}\sum_{({e}_a,{e}_b)\in\bar{S}_{(e_i,e_j)}} [||\tilde{\mathbf{z}^{\mathbb{E}}_{e_i}}-\tilde{\mathbf{z}^{\mathbb{E}}_{e_j}}||_{L2}+\\ & 
    \gamma - 
    ||\tilde{\mathbf{z}^{\mathbb{E}}_{{e_a}}}-\tilde{\mathbf{z}^{\mathbb{E}}_{{e_b}}}||_{L2}]_{+},
\end{aligned}
\end{equation}
where $\gamma $ is a hyper-parameter of margin, $ [x]_+ = max\{0,x\} $ is to ensure non-negative output. $ \bar{S}_{(e_i,e_j)} $ is a collection of negative samples composed of randomly replaced entities $e_i$ and $e_j$ from the seed set $\mathcal{S}$.
\subsubsection{Model training}
We combine three losses to achieve the final training objective of our method:
\begin{equation}
    \mathcal{L} = \mathcal{L}_{ea} + \lambda(\mathcal{L}_{inter} + \mathcal{L}_{intra}),
\end{equation}
where $ \lambda $ is a hyper-parameter to adjust the three loss functions.
\section{Experiment}
In this section, we conduct extensive experiments on four public datasets to demonstrate the superiority of our method. Ablation studies validate the effectiveness of each module. Additionally, visualization of the entity embedding from the two KGs intuitively shows that our method is more beneficial for the EA task. Finally, we analyze the training efficiency of the method.

\subsection{Experiment setting}
\subsubsection{Datasets}
To fully demonstrate the superiority of our method, we select the OpenEA (15K-V1) dataset \citep{Sun2020ABS}, which includes two monolingual datasets: DBpedia-to-Wikidata (D-W-15K) and DBpedia-to-YAGO (D-Y-15K), as well as two cross-lingual datasets: English-to-French (EN-FR-15K) and English-to-German (EN-DE-15K).
The details of these four datasets are provided in Appendix~\ref{sec:appendix}.
The triples in OpenEA dataset consist of URLs, which not only align with the degree distribution of real-world KGs but also facilitate research on structure-based EA methods. 
We follow the data splits in OpenEA \citep{Sun2020ABS}, where 20\% of the alignments are used for training, 10\% for validation, and 70\% for testing. We report the average results of five-fold cross-validation. 
The results for each fold are presented in Appendix~\ref{sec:appendix2}.

\subsubsection{Implement details}

Our experiment conducted with a single NVIDIA 4090 GPU with 24GB of memory. We initialize the trainable parameters with Xavier initialization \citep{Xavier} and optimize the loss using Adam \citep{adam}. Regarding hyper-parameters, the entity dimension is set to 256, and the relation dimension is set to 32. The margin for the alignment loss is set to 1. 
We grid search the best parameters $\lambda$ for the final training object \{0.1, 1, 10, 100, 300, 1000\}. The details are provided in Appendix~\ref{sec:appendix3}.
GATs and HGCNs both utilize a two-layer network. We generate 5 negative samples for each positive sample. 
During inference, we use Cross-domain Similarity Local Scaling \citep{csls} to post-process the cosine similarity matrix, which is employed by default in some recent works \citep{Sun2020ABS,Liu2023RHGNRH}.
We use H@k and MRR as evaluation metrics to assess our method, with higher values indicating a greater number of correctly matched entities. We select $k$ values of $\{1, 5\}$.
\subsubsection{Baseline}
To comprehensively evaluate the superiority of our method, we categorize our baselines into three groups: TransE-based, GNN-based, and related methods.
\begin{itemize}
    \item TransE-based EA. These methods leverage variants of TransE to model each triple individually, utilizing strong local structural information: MTransE \citep{Chen2016MultilingualKG}, IPTransE \citep{Zhu2017IterativeEA}, AlignE \citep{Sun2018BootstrappingEA}, and SEA \citep{Pei2019SemiSupervisedEA}.
    \item GNNs-based EA. These methods aggregate neighborhood information by stacking multiple layers of networks: GCN-Align \citep{Wang2018CrosslingualKG}, AliNet \citep{Sun2019KnowledgeGA}, KE-GCN \cite{ke-gcn}, and RHGN \citep{Liu2023RHGNRH}.
    \item Related methods. We classify these four methods into one category: to our knowledge, HyperKA \citep{Sun2020KnowledgeAW} is the only method for EA in hyperbolic space. GAEA \citep{Xie2023ImprovingKG} uses edge deletion information for graph augmentation. IMEA \citep{Xin2022InformedME} is a strong baseline that combines information from nodes, triples, and relation paths.
    GSEA \citep{gsea} uses singular value decomposition of the adjacency matrix to obtain the global structural information of the entities.
\end{itemize}

\begin{table*}[h]

\setlength{\tabcolsep}{0.5pt}
\begin{tabular*}{\textwidth}{@{\extracolsep\fill}ccccccccccccc}
\toprule%
\multirow{2}{*}{Methods} & \multicolumn{3}{@{}c@{}}{EN-FR-15K} & \multicolumn{3}{@{}c@{}}{EN-DE-15K} & \multicolumn{3}{@{}c@{}}{D-W-15K} & \multicolumn{3}{@{}c@{}}{D-Y-15K} \\\cmidrule(lr){2-4}\cmidrule(lr){5-7}\cmidrule(lr){8-10}\cmidrule(lr){11-13}%
&  H@1  & H@5 & MRR &  H@1  & H@5 & MRR &  H@1  & H@5 & MRR &  H@1  & H@5 & MRR \\
\midrule
MtransE  & .247 & .467 & .351 & .307 & .518 & .407 & .259 & .461 & .354 & .463 & .675 & .559 \\

IPTransE & .169 & .320 & .243 & .350 & .515 & .430 & .232 & .380 & .303 & .313 & .456 & .378 \\

AlignE & .357 & .611 & .473 & .552 & .741 & .638 & .406 & .627 & .506 & .551 & .743 & .636 \\

SEA & .280 & .530 & .397 & .530 & .718 & .617 & .360 & .572 & .458 & .500 & .706 & .591 \\
\hline
GCN-Align & .338 & .589 & .451 & .481 & .679 & .571 & .364 & .580 & .461 & .465 & .626 & .536 \\

AliNet & .364 & .597 & .467 & .604 & .759 & .673 & .440 & .628 & .522 & .559 & .690 & .617 \\
KE-GCN & .408 & .670 & .524 & .658 & .822 & .730 & .519 & .727 & .608 & .560 & .750 & .644 \\

RHGN* & .500 & .739 & .603 & .704 & .859 & .771 & .560 & .753 & .644 & .708 & .831 & .762 \\

\hline

HyperKA & .353 & .630 & .477 & .560 & .780 & .656 & .440 & .686 & .548 & .568 & .777 & .659 \\

IMEA & .458 & .720 & .574 & .639 & .827 & .724 & .527 & .753 & .626 & .639 & .804 & .712 \\

GAEA* & .548 & .783 & .652 & .731 & .887 & .800 & .618 & .802 & .802 & .671 & .802 & .731 \\

GSEA* & .561 & .803 & .669 & .740 & .893 & .807 & .628 & .819 & .713 & .694 & .836 & .758 \\

\hline
UniEA(ours) & $\textbf{.580}$ & $\textbf{.811}$ & $\textbf{.682}$ & $\textbf{.748}$ & $\textbf{.898}$ & $\textbf{.813}$ & $\textbf{.648}$ & $\textbf{.826}$ & $\textbf{.728}$ & $\textbf{.712}$ & $\textbf{.841}$ & $\textbf{.771}$ \\
\bottomrule
\end{tabular*}
\caption{Entity alignment result of OpenEA Datasets. The best result in each column is highlighted in bold. * indicates results reproduced from their source code, while other experimental results are from \citet{Xie2023ImprovingKG}. For fairness in the experimental results, we modified the GAEA code to use CSLS during the inference phase.}\label{tab2}
\end{table*}

\subsection{Main results}

\begin{figure*}[ht]
\centering 

\subfigure[H@1]{
\label{Fig:a1}
\includegraphics[width=.3\linewidth]{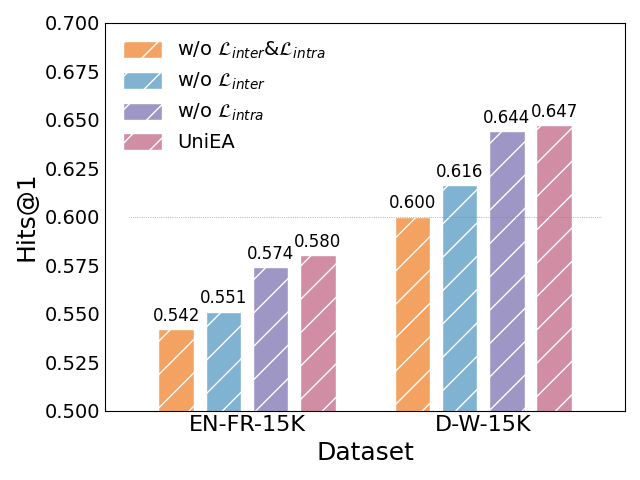}}
\subfigure[H@5]{
\label{Fig:a2}
\includegraphics[width=.3\linewidth]{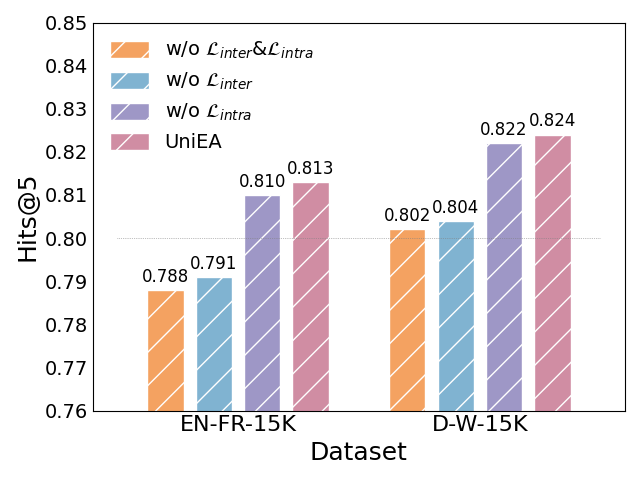}
}
\subfigure[MRR]{
\label{Fig:a3}
\includegraphics[width=.3\linewidth]{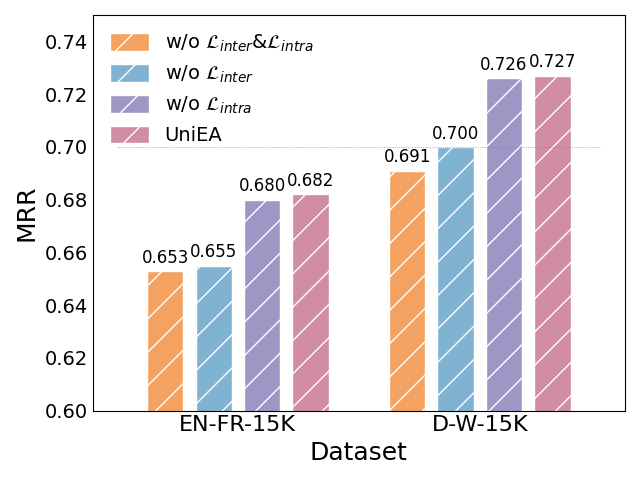}
}
\caption{The results of ablation experiment on EN-FR-15K and D-W-15K.}
\label{fig3}
\end{figure*}
\begin{figure*}[ht]
\centering 

\subfigure[AlignE]{
\label{Fig:a41}
\includegraphics[width=.23\linewidth]{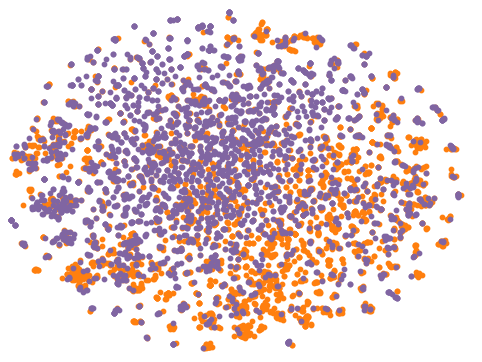}}
\subfigure[GAEA]{
\label{Fig:42}
\includegraphics[width=.23\linewidth]{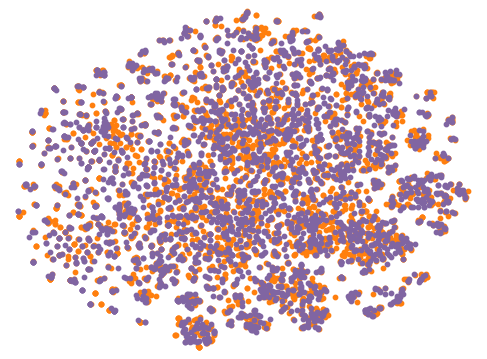}
}
\subfigure[RHGN]{
\label{Fig:43}
\includegraphics[width=.23\linewidth]{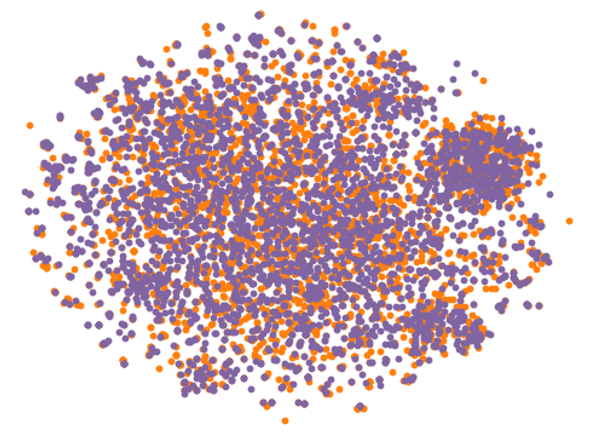}
}
\subfigure[UniEA(ours)]{
\label{Fig:44}
\includegraphics[width=.23\linewidth]{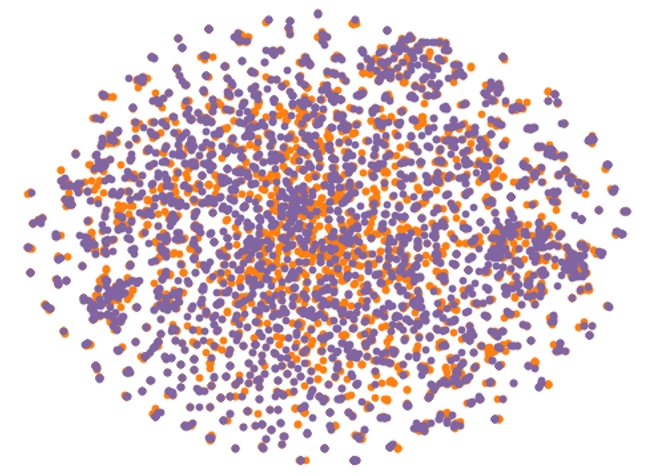}
}
\caption{Visualization of Entity Embedding on EN-FR-15K. Different colors represent different KGs.}
\label{fig4}
\end{figure*}
The results of all methods on OpenEA datasets are shown in Table~\ref{tab2}. Our method outperforms all other methods.
We conducted an analytical comparison with baseline methods and found that IPTransE and SEA are semi-supervised EA methods. Since they cannot effectively learn the structural features of the graph, even with semi-supervised strategies, they fail to improve alignment accuracy.
HyperKA operates in hyperbolic space, but since the hierarchical structure of the four datasets is not particularly pronounced, its performance is inferior to some Euclidean space methods.
Methods that utilize relational semantic information, such as GSEA, IMEA , GAEA, and RHGN, significantly outperform others, indicating that relationships are highly beneficial for entity alignment.
GAEA uses contrastive learning on different views, but its performance is unstable. Our method is not limited by the drawbacks of traditional graph augmentation methods. It not only captures neighborhood information in Euclidean space but also learns the hierarchical structure in hyperbolic space.
Our experiments demonstrate that learning the hierarchical structure in knowledge graphs is crucial.

\subsection{Ablation studies}
We conducted ablation experiments by removing $ \mathcal{L}_{inter} $, $ \mathcal{L}_{intra} $, and $ \mathcal{L}_{inter} \& \mathcal{L}_{intra} $. As shown in Figure~\ref{fig3}, all metrics decreased after the removal of each module, demonstrating the effectiveness of each component.
\begin{itemize}
    \item \textbf{w/o} $\mathcal{L}_{inter}$: By removing the learning in hyperbolic space and using GAT to aggregate neighborhood information and learn relational semantics in Euclidean space, we observed a significant decline in all metrics across the datasets. This indicates that learning both Euclidean and hyperbolic embedding through contrastive learning is effective.
    \item \textbf{w/o} $\mathcal{L}_{intra} $: Removing contrastive learning within the $\mathcal{G}_1$ in Euclidean space also resulted in a decline in the method's performance. Our method shows the most significant decline, indicating that it effectively pushes similar but easily confused entities further apart in the embedding space.
    \item \textbf{w/o} $\mathcal{L}_{inter} \& \mathcal{L}_{intra} $: The decline observed after removing it demonstrates that combining $\mathcal{L}_{intra} $ and $\mathcal{L}_{inter}$ is beneficial for enhancing alignment performance.
\end{itemize}
 
\subsection{Visualization of Entity Embedding}
To more intuitively highlight the performance of our method, we use t-SNE \citep{JMLR:v9:vandermaaten08a} to visualize the entity embedding. In the baseline, we categorized all comparison methods into three groups, and then, we selected the top three methods for comparison. We randomly selected 3,000 entity pairs, and the final embeddings are shown in Figure~\ref{fig4}.

The embedding from AlignE (Figure~\ref{Fig:a41}) are clearly the worst, with one KG's embeddings concentrated in the upper left and the other in the lower right. As a result, during the alignment inference phase, it fails to match the correct entities.
GAEA (Figure~\ref{Fig:42}) shows multiple clusters along the edges, with the left half being sparser than the right. The distribution of entity embedding is uneven, which causes similar entities to be placed too close together, leading to incorrect alignments.
RHGN (Figure~\ref{Fig:43})shows a large cluster in the upper right corner, with uneven distribution and poor embedding results. In contrast, our method (Figure~\ref{Fig:44}) exhibits a uniform distribution without noticeable clustering. Observing the surrounding area, the points of different colors in our method overlap completely, indicating better embedding performance. Therefore, our method achieves results superior to other methods.
\subsection{Auxiliary Experiments}
\subsubsection{Parameter sizes analysis} 
We selected four baseline models for comparison of parameter sizes, measured in millions (M), as shown in Table~\ref{tab:size}. GAEA uses a single GAT network for training, significantly reducing its model complexity. Our method requires different networks for training in both Euclidean and hyperbolic spaces, making our model's parameter size nearly double that of GAEA. However, our parameter size is much smaller than that of IMEA, which uses complex features. Despite this, our method outperforms all structure-based methods.
\begin{table}[htbp] 
    \centering
    \begin{tabular}{l c}
        \toprule
        \textbf{Methods} & \textbf{\#Params (M)} \\
        \midrule
        AliNet & $\sim$16.18M \\
        IMEA & $\sim$20.44M \\
        GAEA & $\sim$8.10M \\
        RHGN & $\sim$8.62M \\
        UniEA(ours) & $\sim$15.86M \\
        \bottomrule
    \end{tabular}
    \caption{Methods parameters comparison}\label{tab:size}
\end{table}

\subsubsection{Efficiency analysis}
\begin{figure}[t]
  \includegraphics[width=\columnwidth]{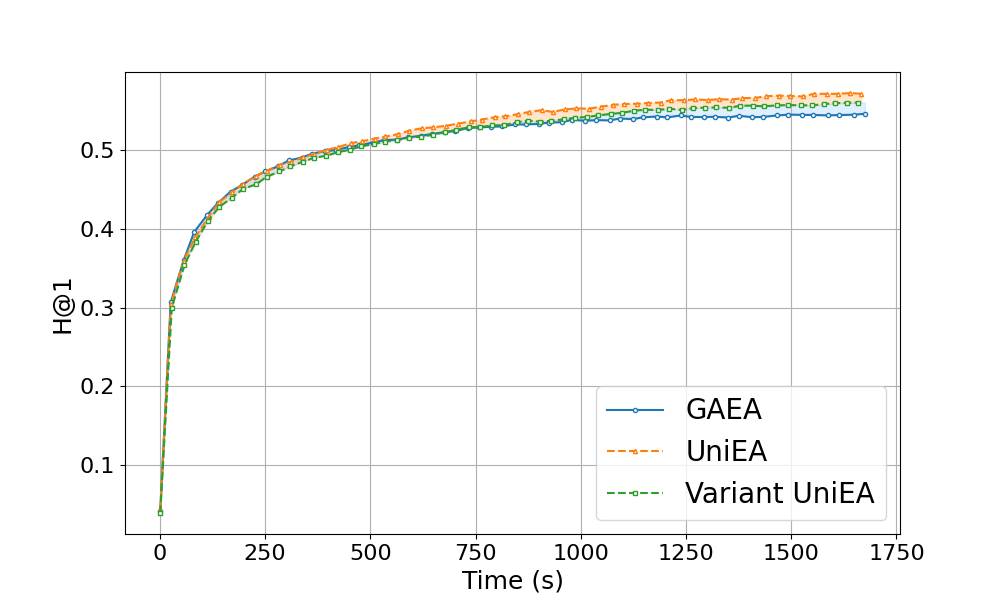}
  \caption{H@1 results and training times on EN-FR-15K.}
  \label{fig:time}
\end{figure}

To evaluate the time efficiency of our method, we conducted a comparative analysis with GAEA. We also included our variant, UniEA-$w/o \enspace \mathcal{L}_{intra}  $, which does not use contrastive learning for similar entities. The final results are shown in Figure ~\ref{fig:time}.
For a fair comparison, we ran 300 epoch and used the same entity embedding dimension of 256 and relation embedding dimension of 32. 

GAEA updates the augmented graph every 10 epochs, which impacts its training efficiency. In contrast, our method does not require augmented graph updates, making it the faster. However, since the UniEA method introduces contrastive learning to push similar entities further apart, it requires longer training time than variant UniEA.
\section{Conclusion}
In this paper, we propose a novel method for EA that unifies dual-space through contrastive learning. We view the learning in hyperbolic space as a specialized form of graph augmentation. Maximizing the consistency of Euclidean and hyperbolic space embeddings using contrast learning Additionally, we employ contrastive learning to increase the distance between embedding of similar entities in Euclidean space, thereby preventing erroneous alignments caused by similarity. Finally, we conduct analyses through ablation studies, visualizations, parameter size comparisons, and evaluations of time efficiency.
\section*{Limitation}
We acknowledge three limitations in our method. First, in the Auxiliary Experiments, we discuss parameter size and time efficiency. While our approach addresses the drawbacks of traditional contrastive learning in generating contrastive views, the introduction of a hyperbolic convolutional network results in a significant number of parameters. Second, our method focuses solely on learning structural and relational information, leaving a wealth of attribute information within KGs untapped. Finally, real-world KGs are predominantly unlabeled, and labeling data is costly. We lack unsupervised or semi-supervised strategies to enhance alignment performance.
\section*{Acknowledgements}
This work is supported by National Natural Science Foundation of China (Nos.62066033);
Inner Mongolia Natural Science Foundation (Nos.2024MS06013, 2022JQ05); 
Inner Mongolia Autonomous Region Science and Technology Programme Project (Nos.2023YFSW0001, 2022YFDZ0059, 2021GG0158);
We also thank all anonymous reviewers for their insightful comments.

\bibliography{custom}

\clearpage
\appendix
\section*{Appendix}
\section{Dataset Statistics} \label{sec:appendix}
Table~\ref{tab1} provides rich information about OpenEA datasets.

\begin{table*}[hbt]
\setlength{\tabcolsep}{1pt}
\begin{tabular*}{\textwidth}{@{\extracolsep\fill}ccccccccc}
\toprule%
\multirow{2}{*}{Dataset} & \multicolumn{2}{@{}c@{}}{EN-FR-15K} & \multicolumn{2}{@{}c@{}}{EN-DE-15K} & \multicolumn{2}{@{}c@{}}{D-W-15K} & \multicolumn{2}{@{}c@{}}{D-Y-15K} \\\cmidrule(lr){2-3}\cmidrule(lr){4-5}\cmidrule(lr){6-7}\cmidrule(lr){8-9}%
& English & French & English & German & DBpedia & Wikidata & DBpedia & YAGO \\
\midrule
\#Ent. & 15,000 & 15,000 & 15,000 & 15,000 & 15,000 & 15,000 & 15,000 & 15,000 \\
\#Rel. & 267 & 210 & 215 & 131 & 248 & 169 & 165 & 28 \\
\#Rel tr. & 47,334 & 40,864 & 47,676 & 50,419 & 38,265 & 42,746 & 30,291 & 26,638 \\
\bottomrule
\end{tabular*}
\caption{The statistics of OpenEA datasets}\label{tab1}
\end{table*}

\begin{table*}[hbt]

\setlength{\tabcolsep}{0.5pt}
\begin{tabular*}{\textwidth}{@{\extracolsep\fill}ccccccccccccc}
\toprule%
\multirow{2}{*}{Fold} & \multicolumn{3}{@{}c@{}}{EN-FR-15K} & \multicolumn{3}{@{}c@{}}{EN-DE-15K} & \multicolumn{3}{@{}c@{}}{D-W-15K} & \multicolumn{3}{@{}c@{}}{D-Y-15K} \\\cmidrule(lr){2-4}\cmidrule(lr){5-7}\cmidrule(lr){8-10}\cmidrule(lr){11-13}%
&  H@1  & H@5 & MRR &  H@1  & H@5 & MRR &  H@1  & H@5 & MRR &  H@1  & H@5 & MRR \\
\midrule

1 & .580 & .807 & .680 & .744 & .901 & .812 & .654 & .831 & .733 & .712 & .838 & .770 \\

2 & .576 & .808 & .679 & .748 & .895 & .813 & .652 & .829 & .731 & .716 & .845 & .774 \\

3 & .585 & .816 & .687 & .752 & .902 & .817 & .646 & .824 & .727 & .717 & .847 & .775 \\

4 & .580 & .810 & .682 & .746 & .895 & .811 & .647 & .824 & .727 & .710 & .838 & .768 \\

5 & .582 & .818 & .685 & .751 & .899 & .815 & .644 & .823 & .723 & .709 & .840 & .769 \\

\hline
UniEA(avg.) & .580 & .811 & .682 & .748 & .898 &.813 & .648 & .826 & .728 & .712 &.841 & .771 \\
\bottomrule
\end{tabular*}
\caption{Entity alignment result of OpenEA datasets on every fold. }\label{tab3}
\end{table*}
\begin{table*}[hbt]

\setlength{\tabcolsep}{0.5pt}
\begin{tabular*}{\textwidth}{@{\extracolsep\fill}ccccccccccccc}
\toprule%
\multirow{2}{*}{$\lambda$} & \multicolumn{3}{@{}c@{}}{EN-FR-15K} & \multicolumn{3}{@{}c@{}}{EN-DE-15K} & \multicolumn{3}{@{}c@{}}{D-W-15K} & \multicolumn{3}{@{}c@{}}{D-Y-15K} \\\cmidrule(lr){2-4}\cmidrule(lr){5-7}\cmidrule(lr){8-10}\cmidrule(lr){11-13}%
&  H@1  & H@5 & MRR &  H@1  & H@5 & MRR &  H@1  & H@5 & MRR &  H@1  & H@5 & MRR \\
\midrule

0.1 & .544 & .787 & .652 & .712 & .883 & .787 & .603 & .806 & .693 & .672 & .824 & .740 \\
1 & .544 & .790 & .653 & .715 & .879 & .788 & .603 & .807 & .694 & .672 & .828 & .742 \\
10 & .547 & .795 & .657 & .718 & .882 & .791 & .616 & .817 & .704 & .688 & .834 & .754 \\
100 & .572 & .811 & .678 & .741 & .895 & .809 & .641 & .830 & .725 & .710 & .838 & .768 \\
300 & .580 & .810 & .682 & .746 & .895 & .811 & .647 & .824 & .727 & .703 & .824 & .759 \\
500 & .577 & .806 & .679 & .744 & .893 & .810 & .641 & .820 & .721 & .690 & .819 & .749 \\
1000 & .572 & .793 & .670 & .742 & .887 & .806 & .634 & .806 & .711 & .682 & .820 & .745 \\
\bottomrule
\end{tabular*}
\caption{Experimental results with different hyper-parameters $\lambda$. }\label{tab4}
\end{table*}

\section{Our methods result of OpenEA Datasets} \label{sec:appendix2}
We used the same parameters for 5-fold cross-validation for each dataset. The reported result UniEA(avg.) in Table~\ref{tab3} is obtained by averaging over five-fold.  

\section{Hyper-parameter settings.} \label{sec:appendix3}
We conducted a single experiment with fold 4, and ultimately chose $\lambda=300$ for the EN-FR-15K, EN-DE-15K, and D-W-15K datasets, while selecting  $\lambda=100$ for the D-Y-15K dataset.
The experimental results are shown in Table~\ref{tab4}.  

\end{document}